\documentclass[letterpaper, 10 pt, journal, twoside]{IEEEtran}
\ifCLASSINFOpdf
    \usepackage{gensymb}
    \usepackage{xcolor}
    \usepackage{amsmath}
    \usepackage{amssymb}
    \usepackage{pifont} 
    \newcommand{\cmark}{\ding{51}} 
    \usepackage[hidelinks]{hyperref}
    \usepackage{cite} 
    \usepackage[linesnumbered,ruled,vlined]{algorithm2e}
    \usepackage{makecell} 

    \usepackage{algpseudocode} 
    \usepackage{booktabs}
    \usepackage{multirow}
    \usepackage{siunitx}
    \usepackage{graphicx}
    \usepackage{threeparttable}
    \usepackage{bbding}
    \usepackage{pifont}
    \usepackage{wasysym}
    \usepackage{amssymb}
    \usepackage{float} 
    \usepackage{stfloats}
    \usepackage{afterpage}
    \usepackage{textcomp}
    
\else
\fi
\begin{document}
%
\title{Semantic Exploration and Dense Mapping of Complex Environments using Ground Robot with Panoramic LiDAR-Camera Fusion}
%
%
%

\author{Xiaoyang Zhan, Shixin Zhou, Qianqian Yang, Yixuan Zhao, Hao Liu, Srinivas Chowdary Ramineni and Kenji Shimada%
\thanks{Manuscript received: May, 13, 2025; Revised August, 4, 2025; Accepted August, 31, 2025.}
\thanks{This paper was recommended for publication by Associate
 Editor L. Heng and Editor S. Behnke upon evaluation of the reviewers’ comments. 
This work was supported by YKK AP Inc. and YKK AP Technologies Lab (NA), Inc. 
\textit{(Corresponding Author: Xiaoyang Zhan)}} 
\thanks{The authors are with the Department of Mechanical Engineering, Carnegie Mellon University, Pittsburgh, PA 15213, USA (e-mail: xzhan2@andrew.cmu.edu).  }%

\thanks{Digital Object Identifier (DOI):10.1109/LRA.2025.3609216}
}
%
%

\markboth{IEEE Robotics and Automation Letters. Preprint Version. Accepted August, 2025}
{Zhan \MakeLowercase{\textit{et al.}}: Semantic Exploration and Dense Mapping of Complex Environments using Ground Robot}

%



\maketitle

\begin{abstract}
This paper presents a system for autonomous semantic exploration and dense semantic target mapping of a complex unknown environment using a ground robot equipped with a panoramic LiDAR-camera system. Existing approaches often struggle to strike a balance between collecting enough high-quality observations from multiple view angles and avoiding unnecessary repetitive traversal. To fill this gap, we propose a complete system that combines mapping and planning. We first redefine the task as completing both geometric coverage and semantic viewpoint observation. Subsequently, we manage semantic and geometric viewpoints separately and propose a novel Priority-driven Decoupled Local Sampler to generate local viewpoint sets. This allows for explicit multi-view semantic inspection and voxel coverage without unnecessary repetition. Building on this, we develop a hierarchical planner that ensures efficient global coverage. In addition, we propose a Safe Aggressive Exploration State Machine, which allows the robot to extend its exploration path planning into unknown areas while ensuring the robot’s safety with recovery behavior and adaptive sampling. Our system includes a modular semantic target mapping component designed to utilize odometry and point clouds from existing SLAM algorithms, enabling point-cloud-level dense semantic target mapping. We validate our approach through extensive experiments in both realistic simulations and complex real-world environments. Simulation results demonstrate that our planner achieves faster exploration and shorter travel distances while guaranteeing a specified number of multi-view inspections. Real-world experiments further confirm the effectiveness of the system in achieving accurate dense semantic object mapping of unstructured environments.

\end{abstract}


%
\IEEEpeerreviewmaketitle

\section{Introduction}
\label{sec: intro}
%
%
%
%

\begin{figure}[h!]
    \centering
    \includegraphics[width=0.95\linewidth]{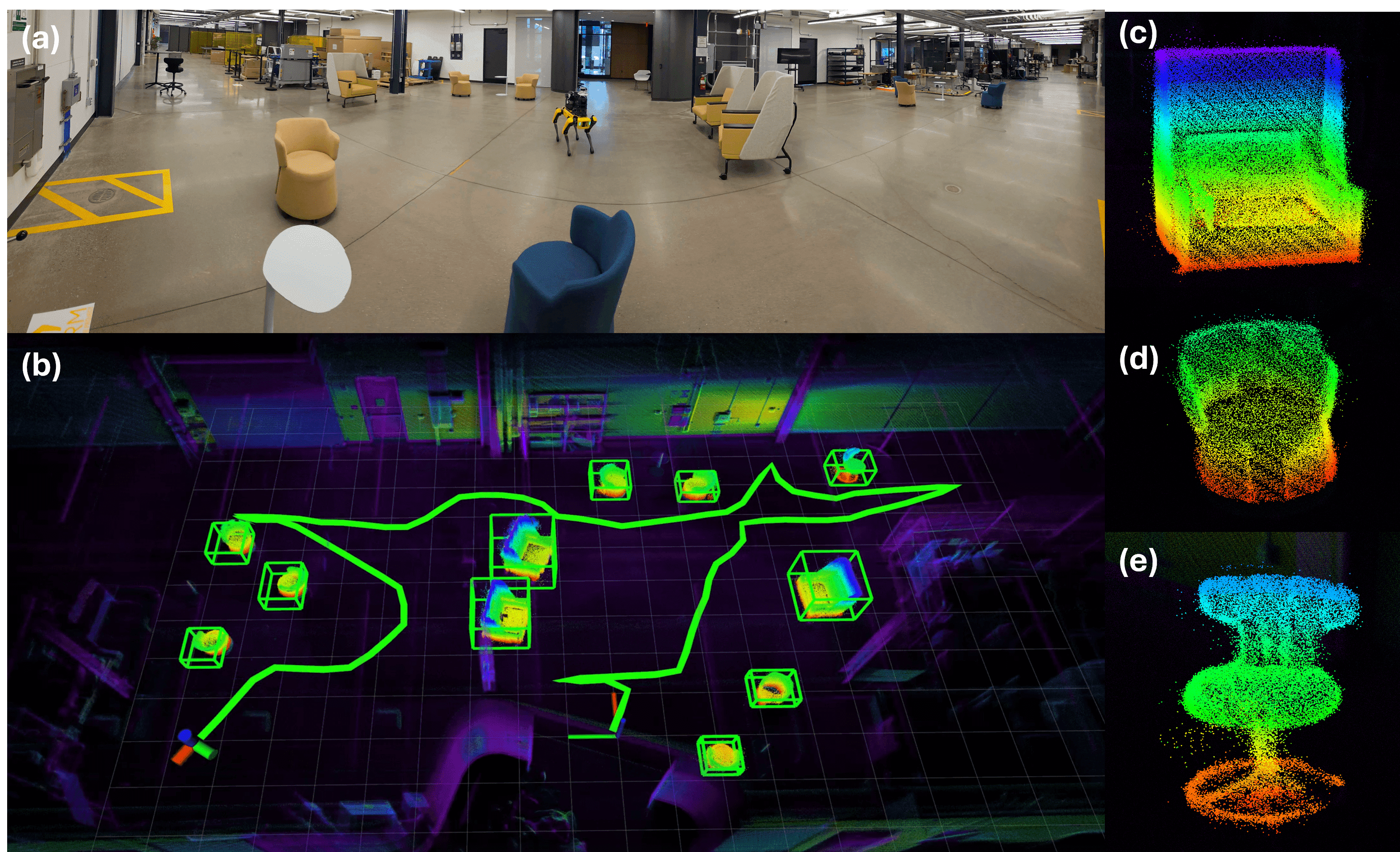}
    \caption{A real-world semantic exploration experiment in a lobby environment. (a) Lobby scene with legged robot. (b) Visualization showing background map, dense object map, and exploration path. Green boxes mark semantic objects (chairs) with highlighted dense maps. (c)-(e) Sample dense target maps. Full results in Secs~\ref{sec: physical_exp}. }
    \label{fig: introduction}
\end{figure}
\IEEEPARstart{W}{ith} the growing success of autonomous exploration algorithms, their applications are drawing increased attention from the research community. Semantic exploration is particularly important in scenarios such as factories and construction sites. Unlike Autonomous Aerial
Vehicles (AAVs) commonly used in prior work, ground robots offer the advantage of accommodating more powerful sensors and computing systems, including high-resolution LiDAR and panoramic camera arrays. This paper focuses on semantic exploration and mapping for ground robots.

Despite advancements, autonomous semantic exploration faces challenges. First, effective integration of mapping and planning is required, as the planner depends on semantic information to make informed decisions. Multi-view observation of target objects can help in many cases. For example, aggregating distinct observations around the object helps mitigate false detection results from the detector, and visual documentation like images from diverse perspectives also helps human supervisors in human-robot interaction scenarios. Therefore, the planner must actively seek informative viewpoints while computing a path for geometric coverage as short as possible. Existing methods either assign semantic weights to potential exploration goals that prioritize positions near target objects \cite{find_things,semantic_search} or sample viewpoints that cover unknown regions \cite{star_searcher}. However, the former often causes repetitive movements around objects, while the latter neglects the importance of complete and diverse observation from multiple view angles. There is still a lack of a planner that strikes the balance for both.

Another difficulty is that robots operating in complicated environments need to balance exploration speed and collision risks. Existing planners usually restrict exploration to known free spaces to guarantee safety, but this over-conservative strategy results in short-sighted decision making and inefficient paths \cite{uav_tunnel}. Although \cite{falcon} extends global path guidance into unknown space, it still does not allow for the search and execution of local-level paths in unknown areas. However, executing aggressive exploration paths in unknown areas increases collision risks, especially at higher speeds, which limits planner performance.

To address the above challenges, this paper proposes a system that performs semantic exploration planning and dense semantic target mapping in an unknown environment. We develop a modular, object-centric semantic target mapping component that operates on odometry and dense point cloud data from existing SLAM systems, without requiring tight integration with SLAM backend. The planning component defines the semantic exploration task as inspecting all unknown voxels and visiting evenly distributed semantic viewpoints to ensure high-quality mapping. To support the integration between mapping and planning, we introduce a hierarchical planner that manages semantic viewpoints and geometric exploration viewpoints in a unified framework. We propose a novel Priority-driven Decoupled Local Sampler to handle the two types of viewpoints separately and generate a minimum set of viewpoints that supports the completeness of both semantic and geometric coverage tasks. The sampler is then integrated with a global planner to generate the global optimal path by solving an asymmetric traveling salesman problem (ATSP). After that, to support aggressive exploration, a novel Safe Aggressive Exploration State Machine is proposed to ensure aggressive and resilient exploration in crowded regions. 

In summary, this work has the following contributions:
\begin{itemize}
    \item \textit{A modular semantic target mapping component based on panoramic LiDAR-camera fusion}, 
    directly leveraging the outputs from existing SLAM systems without requiring integration into back-end optimization.
    \item \textit{A novel Priority-driven Decoupled Local Sampler} that manages the semantic and geometric viewpoints in a decoupled manner and generates a viewpoint set for global path planning.
    \item \textit{A safe aggressive exploration state machine} that supports the robot in proactively selecting viewpoints and executing local paths in unexplored areas, while maintaining safety by actively checking and recovering.
    \item \textit{Real-world tests in construction and industrial environments} that reflect actual application deployment and validate the effectiveness of the complete system.
\end{itemize}

%

\section{RELATED WORK}
Autonomous robot exploration has made great progress over the decades \cite{frontier_classic,fuel,tare,gbp}. One stream of methods is based on frontiers, which are the boundaries between free space and unknown space. Yamauchi \cite{frontier_classic} greedily chooses the closest frontier as the next goal, while later methods \cite{fuel, racer} apply more sophisticated frontier clustering, viewpoint generation, and path planning techniques. Another stream of methods is based on Next-Best-View (NBV) and graph sampling like RRT or PRM. In \cite{nbv}, a graph is built by sampling in safe free space and calculating the information gain for each node. The next best node with the highest average gain is chosen as the next goal. Later, Dang et al. \cite{gbp} proposes rapidly-exploring random graphs for more rapid NBV exploration planning, and other methods 
 \cite{ufoplanner,other_sampling_2,HIRE,ERRT} continue to propose more carefully designed utility functions for information gain and graph structures. In recent years, more algorithms have been formulated as the traveling salesman problem (TSP) \cite{fuel,tare,Lidar_uav_exploration,star_searcher} to improve the efficiency of traversing all viewpoints or graph nodes.

At the same time, the SLAM community has an increasing interest in metric-semantic SLAM. Some of them use sparse representations such as cubes \cite{cube_slam}, ellipsoids \cite{quadrics_slam}, or hybrid geometric shapes \cite{slides_slam}, and combine the parameters into a factor graph to formulate an optimization problem. Others use dense representations such as meshes \cite{kimera} or sufels \cite{suma++}.

With the development of both areas, semantic exploration is gaining more attention \cite{other_se,find_things,star_searcher,vlfm}. Tang et al. \cite{semantic_search} define an object gain function to handle re-observation from multiple resolutions, and Papatheodorou et al. \cite{find_things} propose multiple utility functions to include both background and object information gain to improve the quality of camera-based mesh reconstruction. Tao et al. \cite{active_semantic_slam} introduce semantic loop-closure viewpoints into exploration and formulate a Correlated Orienteering Problem (COP), but this formulation does not guarantee visiting all semantic viewpoints and requires a hard time-constraint estimation.  Luo et al. \cite{star_searcher} further propose a hierarchical planning framework to solve exploration with target search. VLM-based semantic navigation \cite{vlfm, seek} represents a recent sub-branch of exploration research, but recent work generally aim to find
objects as soon as possible without pursuing full coverage
of the environment.
Meanwhile, the DARPA subterranean challenge facilitated semantic exploration research. Lei et al. \cite{early_recall} propose a pipeline to achieve high precision and recall of semantic detection and emphasize the importance of multi-view observation. Best et al. \cite{multi_sensor} utilized multi-sensor fusion during exploration and mapping.

None of these methods, however, explicitly define multi-view observation viewpoints for object mapping in exploration, thus suffering from randomness due to unnecessary re-observation or inadequate observation. In addition, most algorithms are still based on the greedy NBV strategy with semantic weights, lacking careful consideration of global optimality. Our method fills this gap by explicitly defining the semantic viewpoint with a certain density around target objects while enabling efficient traversing considering both geometric coverage and semantic observation. Furthermore, we developed a Safe Aggressive State Machine to support aggressive exploration in complicated real-world environments.


\section{PROBLEM FORMULATION} \label{sec: probformulation}

The goal of this paper is to enable a single ground robot, equipped with a panoramic LiDAR and camera fusion system, to autonomously plan its trajectory within an unknown environment and accurately map both the environment itself and specified target objects of interest. The robot should generate a dense point cloud for both the environment and the identified objects, while optimizing trajectory efficiency to ensure timely completion of the mapping tasks.

\subsection{Robot and Sensor Model}
In this work, we employ the Boston Dynamics Spot as a quadruped mobile platform equipped with a specialized sensor system. The sensor suite includes an Ouster OS-1-128 LiDAR and a panoramic camera system. The LiDAR features a 45\degree vertical field of view (FOV) and a 360\degree horizontal FOV, with 128 vertical channels, operating at a frequency of 10 Hz. The panoramic camera system is the Boston Dynamics CAM Payload, which integrates data from five Sony IMX290 cameras to provide a 360\degree $\times$ 165\degree field of view.
\subsection{Environment Model}
The unknown environment is modeled as a bounded 3D volume \(V \subset \mathbb{R}^3\), discretized into individual voxels \(v\). Each voxel \(v\) is associated with an occupancy probability \(P_o(v)\). Before receiving any sensor measurements, each voxel is initialized as \textit{unknown} with \(P_o(v)=0.5\). Through the accumulation of sensor measurements and ray-casting, voxels are eventually classified as either \textit{occupied} \(V_{occ}\) or \textit{free} \(V_{free}\). Due to the constraints from the environment, robot and sensor configuration, some residual voxels \(V_{res}\) may remain unmapped.

\begin{figure*}[b]
    \centering
    \includegraphics[width=0.95\linewidth]{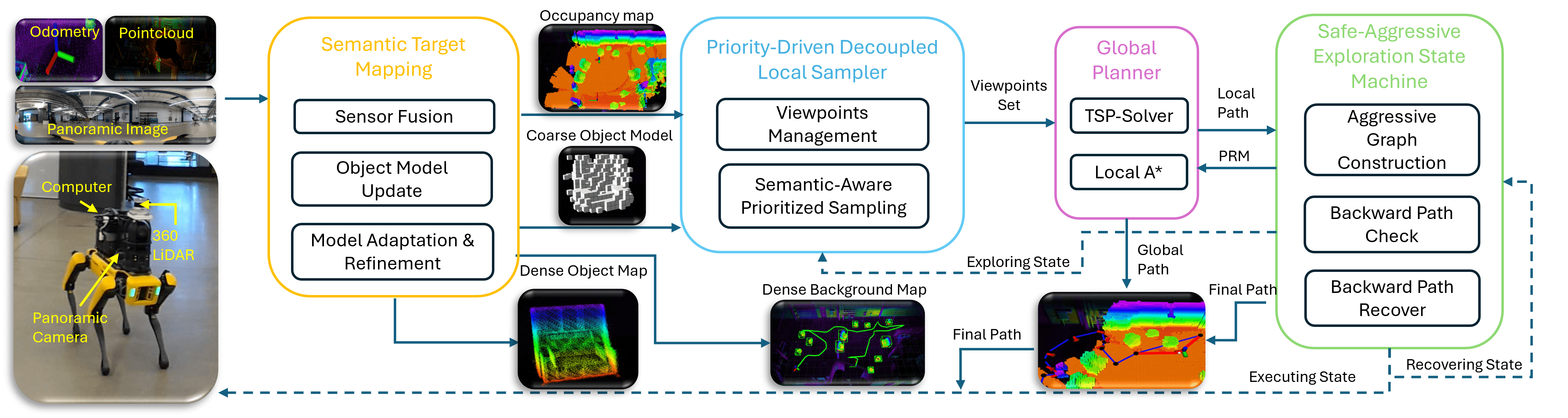}
    \caption{System overview: The system processes odometry, LiDAR, and panoramic image inputs. The mapping module builds object-centric models, occupancy, and dense maps. The planner samples viewpoints based on these maps, creates local viewpoint sets in active regions, and generates optimal paths. The Safe-Aggressive Exploration State Machine maintains aggressive graph construction (AGC) while supporting recovery from pseudo-collision.}
    \label{fig: system_overview}
\end{figure*}

\subsection{Problem Definition} \label{sec: problemdef}
Assume an unknown number \(N\in\mathbf{N}\) of target objects, each associated with a semantic label \(\lambda\) from a semantic label set $\mathbb{L}$. As discussed in Section~\ref{sec: intro}, the semantic exploration task must consider both target object observation and geometric coverage. We divide the space centering around a target object \(O_i\) with radius \(r\) into \(M\) evenly distributed sectors. Each sector \(S^i_m\) is defined as an \textit{observation space}. Then, we define the semantic exploration task as follows:

\textit{\hypertarget{probdef1}{Semantic Exploration}}:
Given the sensor model and configuration $\psi_{robot}$, plan a path $\sigma_{robot}$ starting from an initial configuration $\xi_{robot}$ such that:
\textit{(1)} it covers at least one valid detection in every observation space \(S^i_m\) for any target object \(O_i\), whenever possible;
\textit{(2)} it observes all voxels so that \(V_{free} \cup V_{occ}  = V \setminus V_{res} \); and
\textit{(3)} it extracts all point cloud associated with object $O_i$ as well as the background.

\section{METHODOLOGY}
The proposed system comprises a plug-in semantic mapping module and a hierarchical planner, as shown in Fig. \ref{fig: system_overview}. The mapping module, discussed in Section~\ref{sec: semanticmap}, takes as input the panoramic image $\mathbf{I}$, the LiDAR point cloud $\mathbf{P}$ and the estimation of existing odometry $\mathbf{O}$, subscribes the object bounding box $\mathbf{b_o}$ and the instance mask $\mathbf{m_o}$ from instance segmentation models such as \cite{yolov8}, then outputs the dense point cloud of both the background environment $\mathbf{M_{b}}$ and target objects $\mathbf{M_{o}}$. Concurrently, the hierarchical planner generates the exploration path that ensures both semantic observation and geometric coverage. Section~\ref{sec: localsampler} describes how the Priority-driven Decoupled Sampler decouples semantic observation and geometric coverage and selects geometric and semantic viewpoints. Subsequently, Section~\ref{sec: globalplanner} explains the global path calculation to travel through viewpoints. Finally, a Safe Aggressive Exploration State Machine is proposed in Section~\ref{sec: statemachine} to enable efficient but safe exploration.

\subsection{Sensor Fusion and Pre-processing}
The LiDAR-camera extrinsic calibration is derived from the known SPOT CAM and Ouster LiDAR CAD model. Time synchronization is handled by ROS's approximate time policy with timestamps from the Boston Dynamics and Ouster API. Given the input measurement from both LiDAR and panoramic camera, they are associated by:

\begin{align}
    I_x &= \frac{W_{\text{img}} \left( -\text{atan2}(Y, X) + \pi \right)}{2 \pi} \label{eq:sensor_fusion_1} , \\
    I_y &= f_y \left( -\frac{Z}{\sqrt{X^2 + Y^2}} \right) + y_0 \label{eq:sensor_fusion2} ,
\end{align}
where $W_{\text{img}}$ is the image width, $X, Y, Z$ are the 3D point input from the LiDAR frame, $f_y$ is the vertical focal length of the panoramic camera, and $y_0$ is the vertical coordinate of the panoramic image's principal point, \(I_x\) and \(I_y\) are the associated coordinates in the panoramic image frame.

 Using Eq. \ref{eq:sensor_fusion_1}, \ref{eq:sensor_fusion2}, odometry, object detection, and a raw point cloud are extracted for each instance mask. Then, it is processed by DBSCAN clustering, selecting the closest group to the observation pose, and filtering out points lower than the ground. The instance mask, the associated instance point cloud, and the pose estimation at the current time step are packaged as a measurement $z$.

\subsection{Modular Semantic Target Mapping} \label{sec: semanticmap}
Our plug-in semantic target mapping module builds a coarse-to-fine object-centric voxel model for each target object, updates the model constantly with real-time input measurements, and extracts a dense point cloud from the background SLAM algorithm. It assumes a reasonably accurate odometry estimation. Although severe odometry drift could affect the mapping performance, the increasing availability of robust SLAM frameworks that provide the global pose graph optimization and loop-closure to reduce odometry drift could improve the system in future. Since it is out of scope of this paper, we will not discuss them here.

\subsubsection{Data Association} Upon receiving a processed point cloud, the module matches it to existing objects by bounding box intersection-over-union (IOU). If the best IOU falls below a threshold \(T_{iou}\), a new object model is initialized.

\subsubsection{Model Update}
The object model periodically incorporates new observations and processes multi-view observations statistically, accounting for potential false positives/negatives from varying view angles. Specifically, for each object model $O_i$, observations are grouped into $n$ observation sets. Each set corresponds to the observation space defined in \hyperlink{probdef1}{\textit{Problem Definition}}.
Within each observation set, the model retains a keyframe measurement $z^n_i$ and stores its corresponding mask $m^n_i$ and pose estimation $p^n_i$. The key frame is determined by choosing the best \textit{mask score}:
\begin{equation}
    S_{mask} = IOMax(V_{mask}, V_{model}) * V_{mask},
\end{equation}
where $V_{mask}$ and $V_{model}$ are bounding box volumes of the mask and current model, respectively. IOMax is the intersection over the maximal box. This score prioritizes masks with complete object measurement while avoiding an over-segmented point cloud with an unexpectedly large size.

 However, false positive detection due to a complicated background could also produce a false positive keyframe measurement. Therefore, we model the validity of keyframes as a Bernoulli random variable. The estimation of probability is given by the number of valid detections over the total number of observations in this observation set, where a valid observation means a mask score greater than a threshold $T_{score}$. Only keyframes with a true positive probability greater than $T_{p}$ will be kept.
Each accepted observation is integrated into the voxel model, updating per-voxel probabilities.  After this, for each observation set, the object model is ray-cast back to the key frame mask $m^k_i$, taking occlusions into account. The voxels outside the mask decrease its probability by $p_{dec}$, while those inside the mask increase the probability by $p_{inc}$. The above process is repeated for all observation sets.

\subsubsection{Map Adaptation and Refinement}
Errors in the data association process can lead to two common issues: a single object's measurements being split across different object models, or measurements from two different objects being assigned to the same object model. We propose two strategies to address these issues: model merging and model re-centering.
For the former case, model merging is applied when two object models have an Intersection over Minimum (IoMin) greater than a threshold $T_{IOM}$. In this scenario, their data is merged to form a new object model. The probabilities of belonging to the object model are averaged, and the observation sets and key frames from both models are concatenated.
To address the latter case, we apply model re-centering. During each model update iteration, both the center of geometry and mass are recalculated. If their distance exceeds a threshold $T_{doc}$, the model is re-centered to its center of mass, discarding any information outside the new map boundary. This adjustment helps to "pull back" the model when incorrect measurements have been associated and have biased the map.
Finally, the dense point cloud from LiDAR SLAM will be registered and stored in each voxel of a model. Filtering out voxels with too sparse a point cloud, a dense object map is extracted.

\subsection{Priority-driven Decoupled Local Sampler}\label{sec: localsampler}
\begin{figure}[h!] 
    \centering
    \includegraphics[width=0.98\linewidth]{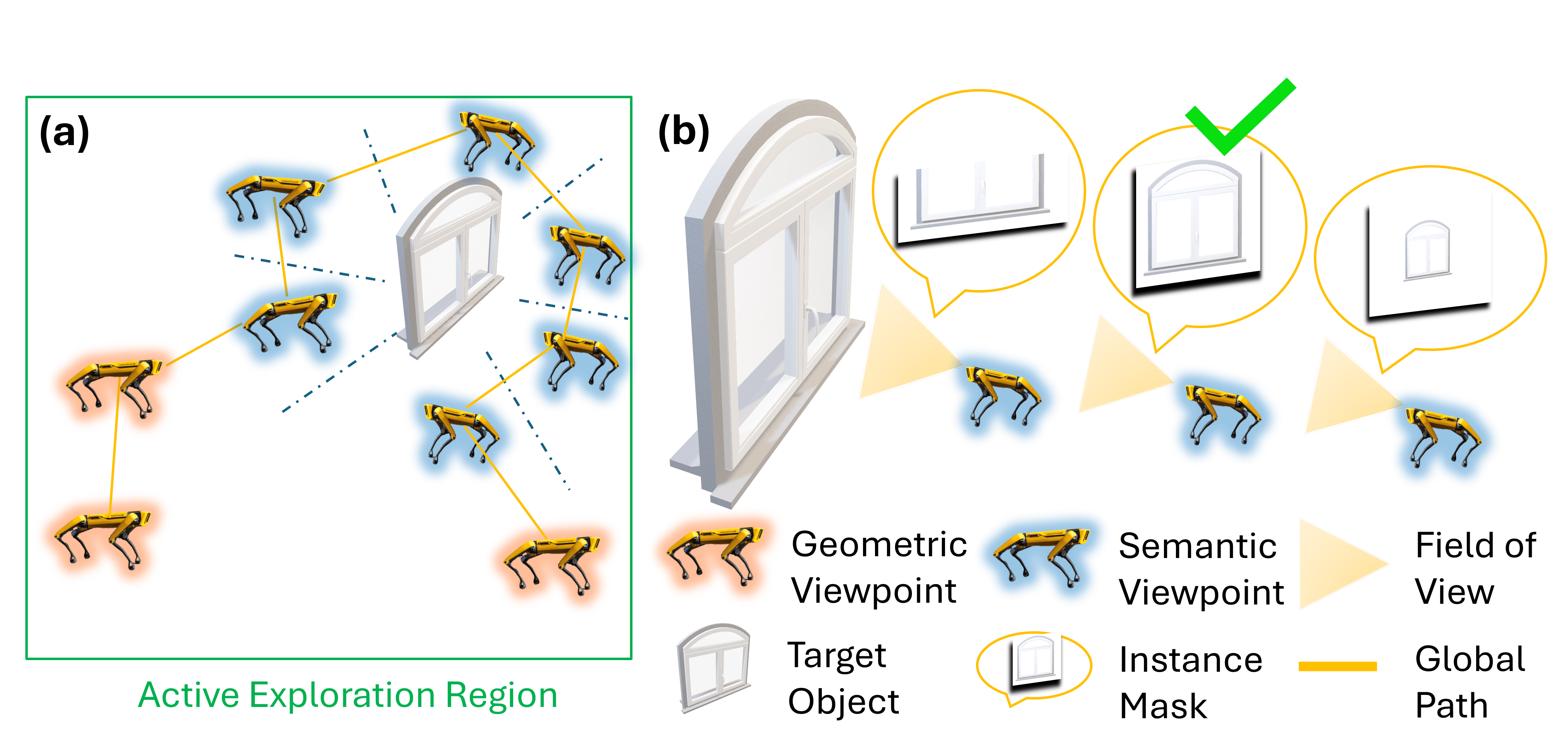}
    \caption{Overview of Priority-driven Decoupled Local Sampler. (a) Sampling semantic and geometric viewpoints in a local active exploration region. (b) Semantic viewpoint sampling along one observation space. The viewpoint with the best observation score ($2^{nd}$ viewpoint from left to right) is selected. Viewpoint with incomplete observation ($1^{st}$ viewpoint) is discarded. }
    \label{fig: viewpoints}
\end{figure}

During each planning iteration, only active exploration regions that contain uncovered frontiers are considered. Among these, the local sampler is invoked only for regions that intersect with the robot’s Local Planning Horizon (LPH), selecting suitable viewpoints for exploration, as illustrated in Fig. \ref{fig: viewpoints}(a) . This section is divided into two parts: the first describes the generation, updating, and management of semantic and geometric viewpoints; the second explains how to construct a set of viewpoints that collectively cover the local exploration region and associated observation spaces.

\subsubsection{Semantic and Geometric Viewpoint Management}As described in Sections \ref{sec: problemdef} and \ref{sec: semanticmap}, the robot must observe targets from all accessible observation spaces. To achieve this, semantic viewpoints are sampled along the centerline of each observation space at a fixed spatial resolution, as shown in Fig. \ref{fig: viewpoints}(b). For each center line, the candidate viewpoint with the highest observation score is selected. The score is calculated by projecting the object model into the camera frame and taking the area of the model’s mask. To ensure observation quality and completeness, two types of viewpoints are discarded: \textit{(a)} those resulting in occluded object observations, and \textit{(b)} those yielding incomplete object observations, determined via ray-casting from the current pose. Fig. \ref{fig: viewpoints}(b) shows an example of viewpoint type (b), where the first
viewpoint from left to right will produce an incomplete observation. The sampler will sample
further viewpoints along the same direction with a fixed resolution interval and pick the one
with the best observation score. Viewpoints are dynamically updated following two rules: (a) if
the robot passes by an observation space and reaches an observation score above a threshold $T_{obs}$, the corresponding viewpoint is removed; and (b) if a viewpoint falls into a voxel
being updated to occupied, it is discarded, and an alternative viewpoint will be selected by
repeating the above process. After updating, the viewpoints are registered in a 2D grid with resolution $g$ and assigned to the corresponding active exploration region. The planner considers only the grid centers as viewpoints to reduce redundant visits to nearby viewpoints when the distribution is crowded.

Geometric viewpoints are obtained using a hierarchical Probabilistic RoadMap (PRM). At the lower level, a dense PRM is constructed to support fine-grained local path planning. At the higher level, nodes are selectively sub-sampled from the dense PRM by maintaining a controlled node density to serve as geometric viewpoints. Only the sampled viewpoints are assigned an information gain (IG) score, defined as the number of unknown voxels visible from its pose.

\subsubsection{Semantic-Aware Prioritized Sampling} \label{sec: semanticsampling}
The sampling of geometric and semantic viewpoints must address two key factors: the potential mutual overlap between viewpoints (known as the submodularity property \cite{tare}) and the need for specific semantic viewpoints for object observation. To account for these, we propose a semantic-prioritized sampling mechanism incorporating viewpoints' mutual overlap into consideration. 
Our approach begins by adding all frontiers within the current exploring region to a frontier queue. To ensure effective multi-view observation, our sampler gives precedence to semantic viewpoints, pushing them to the viewpoints queue and calculating their frontier coverage. The covered frontiers are then removed from the queue. Subsequently, geometric viewpoints receive NBV-reward based on accumulated information gain: 
\begin{equation}
    R_{nbv}^k = \frac{\Sigma_{i=1}^n IG_i}{\Sigma_{i=1}^nT_{i-1,i}},
\end{equation}
\begin{equation}
    T_{i-1,i} = \frac{\arctan(\frac{y_i-y_{i-1}}{x_i-x_{i-1}})}{\omega} + \frac{\|n_i, n_{i-1}\|_2}{v}, n_i \in \sigma_k,
\end{equation}
where $\sigma_k$ is the planned path to viewpoint k, comprising $n$ nodes $n_i$. $IG_i$ is the information gain of $n_i$, $T_{i-1,i}$ is the estimated travel time from $n_{i-1}$ to $n_i$. $\omega$ and $v$ are the expected angular and linear velocity.

 All geometric viewpoints are pushed into an NBV-reward priority queue and are sampled in descending order. Each viewpoint undergoes pseudo-sampling to apply its corresponding frontier coverage. The sampler accepts viewpoints with significant incremental coverage, applying their effects to the frontier queue while dropping those with minimal impact.  The process terminates when either the reward queue is empty or the current sample's reward becomes negligible.

\begin{algorithm}[htbp]
\DontPrintSemicolon
\SetAlgoLined
\KwIn{$\eta\in \{exploring, recovering, executing\}$}

\uIf{$\eta = \text{exploring}$}{
    \textbf{AggressiveGraphConstruction()} \;
    
    $success, \sigma_{path} \leftarrow \textbf{exploreReplan()}$ \;
    
    \uIf{$success$}{
        $\eta \leftarrow \text{executing}$ \;
    }
    \uElse{
        $\eta \leftarrow \text{recovering}$ \;
    }
}
\uElseIf{$\eta = \text{executing}$}{
    $safe \leftarrow \textbf{backwardPathCheck()}$ \; 
    
    \uIf{\textbf{not} $safe$}{
        $\eta \leftarrow \text{exploring}$ \;
    }
}
\uElseIf{$\eta = \text{recovering}$}{
    \textbf{backwardRecover()} \; 
    
    $\eta \leftarrow \text{exploring}$\;
}
\caption{Safe-Aggressive Exploration State Machine}
\label{alg:state_machine}
\end{algorithm}

\begin{figure*}[b]
    \centering
    \includegraphics[width=0.955\linewidth]{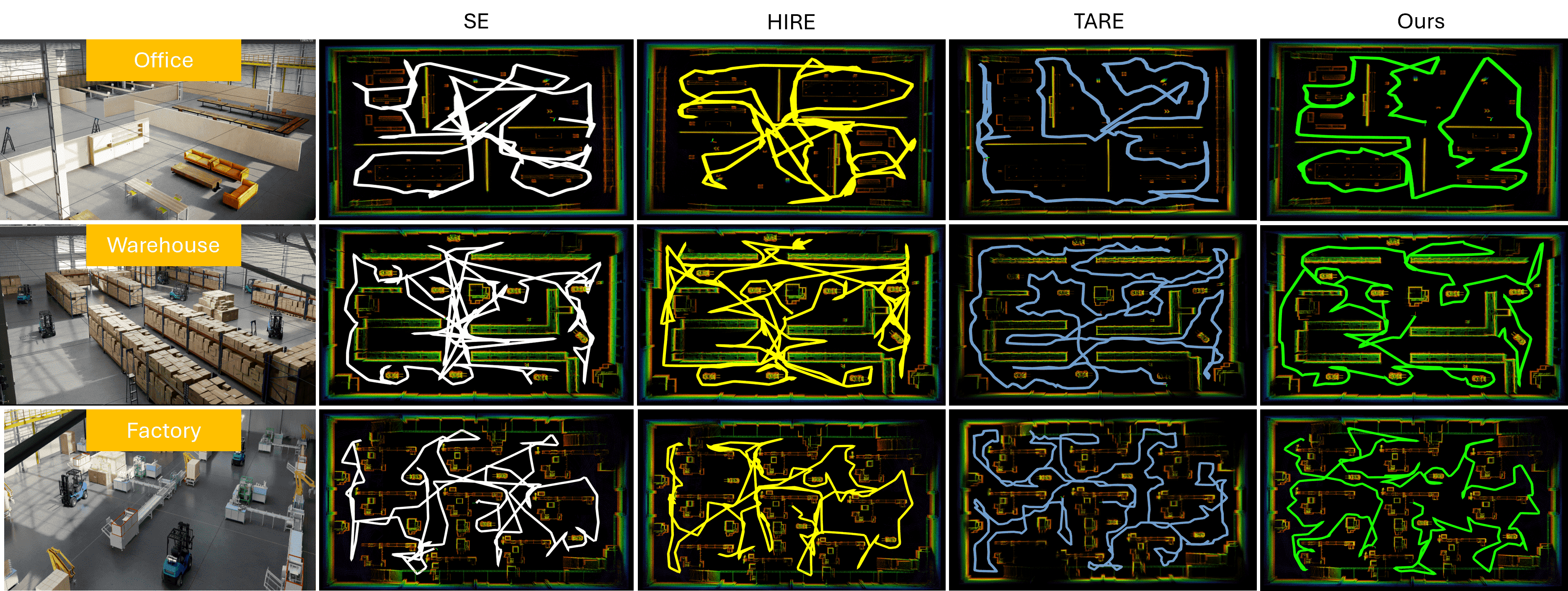}
    \caption{Exploration trajectories of various methods in all three simulator environments. Our method produces more consistent coverage paths with fewer redundant revisits, demonstrating higher exploration efficiency.}
    \label{fig:benchmark_vis}
\end{figure*}

\subsection{Global Planner} \label{sec: globalplanner}
The global planner routes through all active regions containing uncovered frontiers, operating in a receding-horizon approach—executing only one global step before recalculating. We formulate this as an Asymmetric Traveling Salesman Problem (ATSP) similar as \cite{fuel}. Unlike \cite{fuel}, since only the first path segment is executed precisely, the travel time from the current position to the $k^{th}$ viewpoint is computed as:
\begin{equation}
    \mathbf{C_{0,k}} = t^{0,k}_{est} = \Sigma_{i=1}^n T_{i-1,i},
\end{equation}
where $n$ is the number of nodes in the shortest path found by the $A^*$ algorithm. The travel costs among other viewpoints, which may even be located in unknown spaces, are evaluated by Euclidean Distance:
\begin{equation}
    \mathbf{C_{i,j}} = t^{i,j}_{est} = c * \frac{\|VP_i, VP_j\|_2}{v}, i \neq 0,  j \neq 0
\end{equation}
where $VP$ is the viewpoint, and c is a time-cost coefficient.

The planner keeps recomputing in a receding-horizon manner until all viewpoints in active exploration regions are covered, marking the end of exploration.

\subsection{Safe Aggressive Exploration State Machine} \label{sec: statemachine}
Traditional planners confine paths to free space for safety, resulting in short-sighted goal selection. Conversely, naively executing paths that are partially unknown risks instability through overshooting or "pseudo-collisions" (false collision perceptions that cause unexpected halts). We propose an Aggressive Exploration State Machine with three states:$\{exploring, executing, recovering\}$, explicitly extending local path execution into unknown areas while recovering safely from "pseudo-collisions".

\textit{(1)} In the $exploring$ state, the planner aggressively constructs and dynamically maintains a locally safe PRM graph by Aggressive Graph Construction (AGC). It samples nodes in both unknown and free space, assuming all non-occupied connecting edges are valid. The planner performs A* path search and shortcuts using the same assumption.
We adopt an adaptive PRM sampling strategy following \cite{eth_navigation} that prioritizes areas with fewer nodes. As the robot moves, the planner dynamically updates the dense PRM by removing newly occupied nodes with associated edges inside the LPH at a certain frequency. Unlike \cite{HIRE}, our approach dynamically updates the PRM to account for mistaken assumptions about unknown space, maintaining a locally safe PRM during aggressive sampling.
Upon finding a valid exploration path, the system transitions to the $executing$ state. If no path exists, the system identifies a "pseudo collision" and enters the $recovering$ state.

\textit{(2)} During $executing$ state, the system conservatively verifies whether the path and its surrounding areas (within a safe distance) consist of actual free space, checking in backward order. If deemed unsafe, the system reverts to the exploring state to generate a new path.

\textit{(3)} In the $recovering$ state, we propose a backward recovery mechanism to guide the robot away from the pseudo-collision point and resume exploration. Once the recovery state is activated, the robot backtracks along its last valid exploration path. To ensure that it retreats to a sufficiently distant location, the robot iteratively extends the recovery trajectory backward until it reaches a predefined threshold $T_{\text{back}}$. Simultaneously, the $exploring$ state is reactivated in parallel to proactively plan a new exploration path.

\section{Experiments and Results}
\subsection{Implementation Details}
We configure exploration parameters based on environment scale: 15 m grid size with 8 m LiDAR effective range for large-scale simulations, 5 m grid and 3.5 m range for construction sites, and 10 m grid with 5.0 m range for lobby scenarios. We use a safe distance of 0.1 m as path planning shortcut tolerance. All setups use N=8 viewpoints, 0.1 m resolution for occupancy maps and coarse object models, and \text{FAST LIO2} \cite{fastlio} for real-time background SLAM.
\subsection{Benchmark Comparisons}
To evaluate the performance of our hierarchical planner, we performed benchmark experiments in three realistic industrial environments in Isaac Sim: \textit{1)} Office (40$\times$40$\times$5 m) with 13 chairs as target objects, Warehouse (40$\times$60$\times$5 m) with 9 forklifts as target objects and Factory (45$\times$70$\times$5 m) with 11 robot arms and forklifts as target objects. To demonstrate cross-platform applicability, we used a Husky UGV with a velocity range of 0.5 m/s to 2.2 m/s. The robot is equipped with one LiDAR and two cameras (front and rear) to simulate 360° sensing. To isolate planning performance, we assume ground-truth 3D object detections.

\begin{table}[htbp]
    \centering
    \renewcommand{\arraystretch}{1.2}
    \begin{threeparttable}
        \caption{Benchmarking exploration performance in various scenes}
        \label{tab:comparison_final_spaced}
        \scriptsize  
        \setlength{\tabcolsep}{1.5pt}  
        
        \begin{tabular}{
            l
            l
            c
            c
            c
            c
        }
            \toprule
            \textbf{Scene} & \textbf{Method} & 
            {\thead{Runtime\\(s)}} & 
            {\thead{Path Length\\(m)}} & 
            {\thead{Map Comp.\tnote{a}\\(\%)}} & 
            {\thead{VP Comp.\tnote{b}\\(\%)}} \\
            \midrule

            \multirow{4}{*}{\textbf{Office}} 
                & SE \cite{find_things} & $536.2\pm8.1$ & $370.7 \pm 4.6$ & $94.7$ & $99.6$ \\
                & HIRE \cite{HIRE} & $613.7\pm59.5$ & $380.6 \pm 34.6$ & $\mathbf{96.3}$\cmark & $97.3$ \\
                & TARE \cite{tare} & $598.0\pm55.7$ & $340.6 \pm 34.4$ & $94.8$ & $66.0$ \\
                & \textbf{Ours} &$\mathbf{326.1\pm11.3}$\cmark & $\mathbf{197.6 \pm 6.2}$\cmark & $95.3$ & $\mathbf{100.0}$\cmark\\
            \midrule 

            \multirow{4}{*}{\textbf{Warehouse}} 
                & SE \cite{find_things} & $1092.0 \pm 72.7$ & $738.6 \pm 58.3$ & $93.3$ & $97.1$ \\
                & HIRE \cite{HIRE} & $1287.0 \pm 194.9$ & $1013.2 \pm 144.9$ & $94.2$ & $\mathbf{100.0}$\cmark \\
                & TARE \cite{tare} & $863.0 \pm 48.6$ & $485.1 \pm 23.6$ & $92.1$ & $82.5$ \\
                & \textbf{Ours} & $\mathbf{690.5 \pm 30.3}$\cmark & $\mathbf{444.1 \pm 20.3}$\cmark & $\mathbf{94.6}$\cmark & $\mathbf{100.0}$\cmark \\
            \midrule

            \multirow{4}{*}{\textbf{Factory}} 
                & SE \cite{find_things} & $1110.9 \pm 119.8$ & $693.5 \pm 108.8$ & $94.8$ & $97.5$ \\
                & HIRE \cite{HIRE} & $1247.1 \pm 215.2$ & $858.4 \pm 154.0$ & $\mathbf{96.4}$\cmark & $98.1$ \\
                & TARE \cite{tare} & $900.6 \pm 45.8$ & $530.8 \pm 30.5$ & $90.8$ & $76.4$ \\
                & \textbf{Ours} & $\mathbf{840.2 \pm 32.8}$\cmark & $\mathbf{519.5 \pm 55.2}$\cmark & $95.9$ & $\mathbf{100.0}$\cmark \\
            \bottomrule
        \end{tabular}

        \begin{tablenotes}[para,flushleft]
            \scriptsize
            \item[a] Percentage of voxel map coverage.
            \\
            \item[b] Percentage of observed semantic viewpoints over total expected number.
        \end{tablenotes}
    \end{threeparttable}
\end{table}

We compared our approach with HIRE, a recent fast exploration algorithm \cite{HIRE}, and Semantic Eight (SE) \cite{find_things}, a recent semantic exploration algorithm\footnote{\scriptsize Since Semantic Eight was originally designed for depth cameras and tightly integrated with a camera-based SLAM system, we re-implemented its planner to leverage our LiDAR-based sensor configuration.}. To ensure fairness and adapt their metrics to align with our task definition, we enhanced HIRE with a semantic weighting scheme where information gain is calculated as the weighted sum of spatial IG and the semantic gain derived from unvisited visible voxels. In addition, we choose TARE \cite{tare}, a well-known ground robot exploration algorithm, as a baseline to validate our performance in geometric exploration. 
Each experiment was repeated five times in each environment. As detailed in Table~\ref{tab:comparison_final_spaced}, our planner consistently outperformed all of these methods, achieving faster exploration times, shorter path lengths, and reduced standard deviations, indicating greater robustness.
The trajectory visualizations in Fig. \ref{fig:benchmark_vis} confirm that our method shows a much more consistent and efficient trajectory to finish the coverage across all three scenes.

\begin{figure}[htbp]
    \centering
    \includegraphics[width=0.93\linewidth]{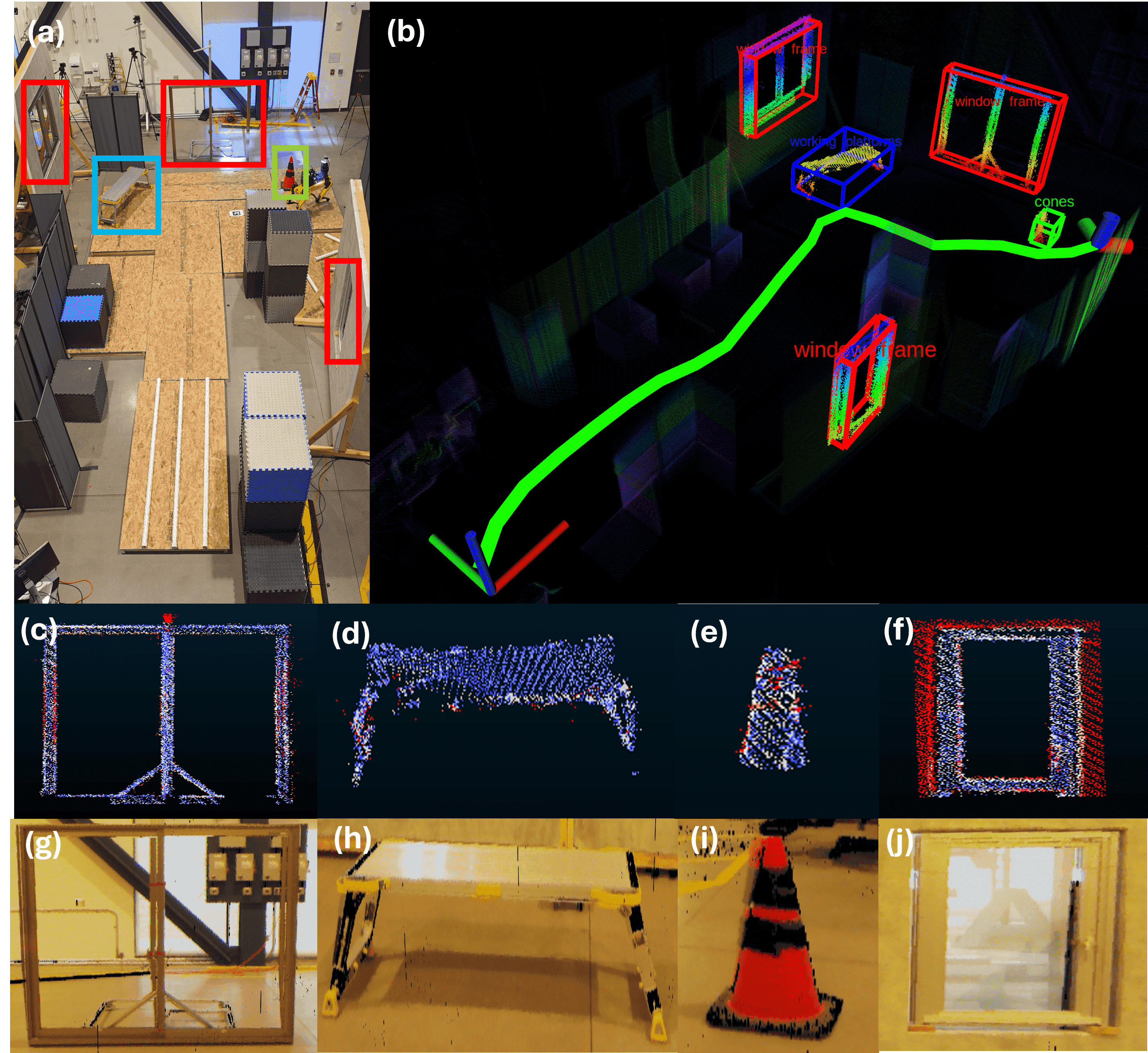}
    \caption{Overview of the construction site scene and mapping results. (a) Bird’s-eye view of the scene with labeled semantic targets. (b) Final map and robot trajectory. (c)-(f) Dense object map with error heatmaps (blue: low error, red: high error). (g)–(j) Ground-truth target object point cloud. Red boxes: window frames; blue: working platform; green: traffic cone. }
    \label{fig: construction_map_vis}
\end{figure}

\subsection{Ablation Study}
We conducted an ablation study in three environments to evaluate the effectiveness of our Priority-driven Decoupled Local Sampler (PDLS) and Safe Aggressive Exploration State Machine (SAESM).

To validate PDLS, we conducted comparisons against a local greedy-information-gain sampler (LGS). It samples the viewpoint with the largest gain greedily in the local active exploration region while keeping all other modules the same as ours\footnote{ LGS strategy generates longer paths and the robot tends to move at a higher speed compared with PDLS. To ensure the completion of the exploration task, we use a safe distance of 0.3 m for local-greedy sampling.}. To incorporate the semantic coverage, we define the overall gain as the weighted sum of the geometric gain and semantic gain derived from unvisited visible voxels of semantic objects. Comparing the Proposed (PDLS) with LGS in Table. \ref{tab:ablation_study_revised}, we achieve higher semantic and geometric coverage rates.

The SAESM consists of two key components: Aggressive Graph Construction (AGC) and the Safe State Machine (SM). We evaluated the effectiveness of each. As shown in Table~\ref{tab:ablation_study_revised}, comparing No-AGC and Proposed demonstrates that AGC significantly reduces the exploration time in most scenarios, confirming its contribution to improved planner efficiency. In the Factory scene, the proposed strategy produces a slightly longer time but shorter path length. This occurs because Factory is so crowded and complicated that aggressive goal selection triggers a lot of stops and recoveries, costing more time than conservative behavior. Furthermore, a comparison between No-SM and Proposed shows that using AGC without SM results in task failures during cluttered environments, underscoring the critical safety role of the SM component.

\begin{table}[htbp]
    \centering
    \renewcommand{\arraystretch}{1.25}
    \begin{threeparttable}
        \caption{Ablation study}
        \label{tab:ablation_study_revised}
        \scriptsize  
        \setlength{\tabcolsep}{1pt}  
        
        \begin{tabular}{
            l
            l
            c
            c
            c
            c
        }
            \toprule
            \textbf{Scene} & \textbf{Method} & 
            {\thead{Runtime\\(s)}} & 
            {\thead{Path Length\\(m)}} & 
            {\thead{Map Comp.\\(\%)}} & 
            {\thead{VP Comp.\\(\%)\tnote{a}}} \\
            \midrule

            \multirow{4}{*}{\textbf{Office}} 
                & No-SM & $175.8^{\dagger}$ & $89.1^{\dagger}$ & $52.8^{\dagger}$ & $50.8^{\dagger}$ \\
                & No-AGC & $433.7 \pm 26.9$ & $270.7 \pm 30.0$ & $\mathbf{96.1}$\cmark & $\mathbf{100.0}$\cmark \\
                & LGS & $496.4 \pm 78.7$ & $272.5 \pm 26.1$ & $94.0$ & $\mathbf{100.0}$\cmark \\
                & \textbf{Proposed} & $\mathbf{326.1 \pm 11.3}$\cmark & $\mathbf{197.6 \pm 6.2}$\cmark & $95.3$ & $\mathbf{100.0}$\cmark \\
            \midrule 

            \multirow{4}{*}{\textbf{Warehouse}} 
                & No-SM & $343.1^{\dagger}$ & $219.2^{\dagger}$ & $66.5^{\dagger}$ & $55.1^{\dagger}$ \\
                & No-AGC & $742.2 \pm 77.9$ & $469.2 \pm 53.7$ & $93.6$ & $\mathbf{100.0}$\cmark \\
                & LGS & $874.6 \pm 205.0$ & $503.6 \pm 47.7$ & $91.6$ & $99.4$ \\
                & \textbf{Proposed} & $\mathbf{690.5 \pm 30.3}$\cmark & $\mathbf{444.1 \pm 20.3}$\cmark & $\mathbf{94.6}$\cmark & $\mathbf{100.0}$\cmark \\
            \midrule

            \multirow{4}{*}{\textbf{Factory}} 
                & No-SM & $327.2^{\dagger}$ & $205.2^{\dagger}$ & $52.3^{\dagger}$ & $34.4^{\dagger}$ \\
                & No-AGC & $840.2 \pm 32.8$ & $542.8 \pm 43.9$ & $\mathbf{96.4}$\cmark & $\mathbf{100.0}$\cmark \\
                & LGS & $946.3 \pm 125.8$ & $585.8 \pm 56.5$ & $93.2$ & $99.3$ \\
                & \textbf{Proposed} & $\mathbf{813.5 \pm 28.5}$\cmark & $\mathbf{519.5 \pm 55.2}$\cmark & $95.9$ & $\mathbf{100.0}$\cmark \\
            \bottomrule
        \end{tabular}

        \begin{tablenotes}[para,flushleft]
            \scriptsize
            \item[$\dagger$] presents the data from incomplete runs; the system failed in all trials.
        \end{tablenotes}
        
    \end{threeparttable}
\end{table}

\subsection{Physical Experiments} \label{sec: physical_exp}
We conducted real-world experiments in two environments—a mock construction site and a workspace lobby—using the Boston Dynamics SPOT robot, with the hardware configuration detailed in Section~\ref{sec: probformulation}. A ROG NUC mini PC equipped with an NVIDIA RTX 4070 GPU was mounted on the robot to perform all the computations.

\begin{figure}[htbp]
    \centering
    \includegraphics[width=0.93\linewidth]{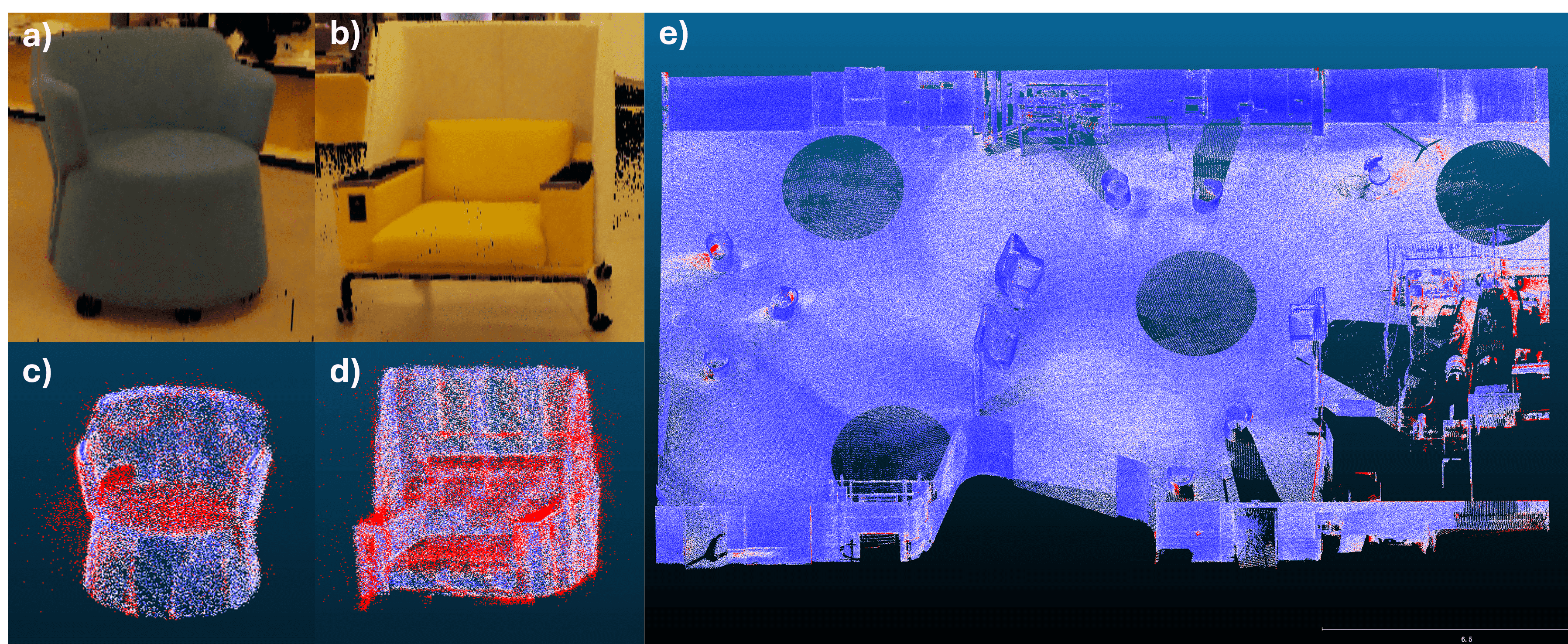}
    \caption{Map and error visualization for the lobby scene. (a)-(b) Ground-truth point clouds of two chairs. (c)-(d) Reconstructed point cloud with error heatmaps. (e) Scene-wide error heatmap comparing. The error increases from blue to red.}
    \label{fig: lobby_error_compare}
\end{figure}

In the 12 m × 8 m mock construction site with five semantic targets (three window frames, one work platform, and one cone), we used YOLOv7 pre-trained on construction data \cite{yolo_construction} for object detection and SAM2 for segmentation. The robot navigated at an average speed of 0.7 m/s (linear) and 0.4 rad/s (angular), completing the exploration in 54 seconds. Fig. \ref{fig: construction_map_vis} provides the visualization of the exploration trajectory, complete map, and error heatmap. The system achieved 87.19\% object map completeness (points within 2.5 cm of the ground truth) with an average error of 3.11 cm, and 99.19\% background map completeness with a 2.27 cm error. In the 12 m × 23 m lobby with 11 chairs of diverse types as semantic targets, we used YOLOV8 instance segmentation pretrained on COCO dataset \cite{yolov8} to provide object detection and segmentation. The exploration was completed in 132 seconds at 0.9 m/s (linear) and 0.6 rad/s (angular). As illustrated in Fig. \ref{fig: introduction}, the robot followed a consistent trajectory that efficiently covered the environment and ensured the observation of all the viewpoints of the targets. Fig. \ref{fig: lobby_error_compare} presents the reconstructed objects and the error analysis. The object map achieved 97.28\% completeness with an average error of 1.75 cm, while the background map reached 99.33\% completeness and 2.31 cm average error. On average, the memory cost per occupied voxel to store a dense cloud is 4,179.70 bytes, and the total memory cost increases linearly with time. Given the 64 GB RAM of our device, there is ample headroom for a larger-scale environment.

\section{Conclusion}
This paper presents a complete system for autonomous semantic exploration and dense target mapping. We propose a hierarchical planner that explicitly handles semantic and geometric viewpoints, a safe-aggressive exploration state machine for efficient coverage, and a modular mapping component that leverages off-the-shelf SLAM for accurate semantic mapping. Extensive simulations validate planning efficiency, while real-world deployments demonstrate precise and complete mapping with minimal exploration effort.


%






%
\bibliographystyle{IEEEtran}
\bibliography{bibliography.bib}



%








\end{document}